\documentclass{article}

\usepackage{xcolor}
\usepackage[final]{corl_2026}
\usepackage{authblk}
\usepackage{graphicx}
\usepackage{float}
\IfFileExists{latexml.sty}{\usepackage{latexml}}{\newif\iflatexml}
\usepackage{flafter}
\usepackage{wrapfig}
\usepackage{amsmath}
\usepackage{amssymb}
\usepackage[justification=justified,singlelinecheck=false]{caption}
\usepackage{booktabs}
\renewcommand{\floatpagefraction}{0.75}
\title{FR-VLA: Implicit Force Reasoning VLA for Contact-Rich Manipulation}

\renewcommand{\Authfont}{\normalsize\bfseries}

\author[1,2]{Jiaju~He}
\author[1,2]{Zhiyuan~Yang}
\author[1,2]{Ruyang~Han}
\author[1,2]{Jiahong~Tang}
\author[1]{Zhiping~Lin}
\author[2]{Haiyue~Zhu\thanks{Corresponding author: \href{mailto:zhu_haiyue@a-star.edu.sg}{\nolinkurl{zhu_haiyue@a-star.edu.sg}}.}}
\affil[1]{School of Electrical and Electronic Engineering, Nanyang Technological University (NTU), Singapore}
\affil[2]{Singapore Institute of Manufacturing Technology (SIMTech), Agency for Science, Technology and Research (A*STAR), Singapore}
\makeatletter
\def\AB@authlist{%
  \makebox[0.30\textwidth][c]{\Authfont Jiaju~He\textsuperscript{\normalfont 1,2}}\hfill
  \makebox[0.30\textwidth][c]{\Authfont Zhiyuan~Yang\textsuperscript{\normalfont 1,2}}\hfill
  \makebox[0.30\textwidth][c]{\Authfont Ruyang~Han\textsuperscript{\normalfont 1,2}}\\[3pt]
  \makebox[0.30\textwidth][c]{\Authfont Jiahong~Tang\textsuperscript{\normalfont 1,2}}\hfill
  \makebox[0.30\textwidth][c]{\Authfont Zhiping~Lin\textsuperscript{\normalfont 1}}\hfill
  \makebox[0.30\textwidth][c]{\Authfont Haiyue~Zhu\textsuperscript{\normalfont 2}\thanks{Corresponding author: \href{mailto:zhu_haiyue@a-star.edu.sg}{\nolinkurl{zhu_haiyue@a-star.edu.sg}}.}}%
}
\makeatother
\hypersetup{
  pdftitle={FR-VLA: Implicit Force Reasoning VLA for Contact-Rich Manipulation},
  pdfauthor={Jiaju He, Zhiyuan Yang, Ruyang Han, Jiahong Tang, Zhiping Lin, Haiyue Zhu}
}

\begin{document}
\maketitle
\vspace{-18pt}
{\centering\normalfont{}Project page: \url{https://fr-vla.pages.dev/}\par}
\vspace{4pt}

\begin{abstract}
Integrating force awareness into Vision-Language-Action (VLA) models involves not merely encoding wrench measurements as an additional modality, but determining when and how strongly physical interaction should influence robot action generation.
Existing VLAs that use force typically inject it through a fixed pathway, without explicitly determining when it is relevant to the current action.
This calls for an intermediate force relevance mechanism inside the VLA, in which visual-language context estimates the relevance of physical feedback to the current task before force enters the action pathway.
FR-VLA addresses this gap with a Vision-Language Force-Reasoning Model that uses learnable force query tokens to derive a force admission signal grounded in visual-language context, and uses this signal to modulate an external tokenized force pathway for action generation.
In real-world experiments on five contact-rich tasks across two robotic platforms, FR-VLA achieves the highest mean success rate over six evaluated stages, and trajectory visualizations show that its force admission varies with the interaction phase.
\end{abstract}


\section{Introduction}

Modern VLAs have become effective generalist policies by grounding robot action generation in visual observations and language instructions through pretrained VLM backbones~\citep{kim2024openvla,black2024pi_0,brohan2023rt2,pi2025pi05}.
However, contact-rich manipulation exposes a limitation of this visual-language foundation: interaction states in insertion, assembly, wiping, and pressing are often only partially observable from images, making force feedback essential for precision-oriented tasks~\citep{leesense2019,gao2022seehearfeel,gelfusion2025,huang2024vitac,huang2025vtrefine}.
The importance of force feedback for action generation varies across task phases, depending on the intended interaction and the evolving contact state~\citep{foar2024,forcematched2024,zhang2021impedanceirl}.
This motivates policies that use visual-language context to determine when and to what extent force feedback should influence action generation.
Consequently, force integration is not merely a problem of sensing or encoding wrench measurements, but one of dynamically grounding physical feedback in visual-language task context~\citep{hao2025tla}.

\begin{figure}[t]
  \centering
  \includegraphics[width=\linewidth]{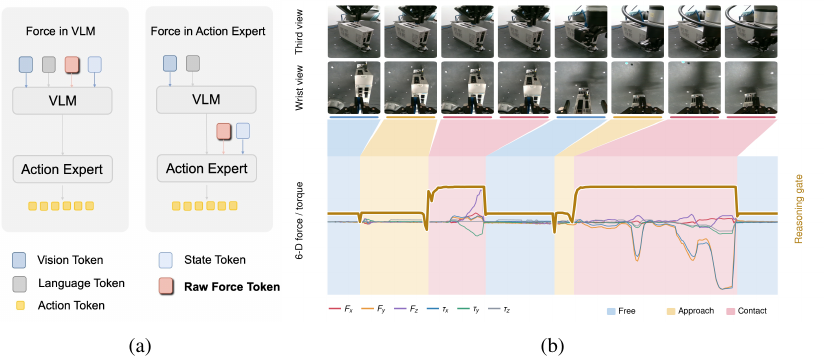}
  \caption{
    {(a)} Two representative force-aware VLA designs either inject force into the VLM token stream or condition the action expert directly.
    {(b)} A Switch Cable Connection trajectory shows synchronized camera observations, six-dimensional wrench measurements, and the force admission signal across task stages.}
  \label{fig:teaser}
\end{figure}

Existing force-aware VLA architectures mainly differ in the point at which force is injected, as illustrated in Figure~\ref{fig:teaser}(a).
Injecting low-dimensional wrench signals into the VLM token stream risks diluting force information within high-dimensional visual-language representations~\citep{forcevla2_2026,fdvla2026,tactilevla2025,vlatouch2025}.
Relocating force conditioning to the action expert avoids this representational imbalance, but bypasses the VLM state where task progress and interaction phase are inferred~\citep{forcevla2025,zhang2025tavla,favla2026,forcepolicy2026}.
Both choices therefore provide force exposure rather than force grounding: the policy can react to measured wrench changes, but it is not explicitly supervised to decide when force should affect action generation under a task-dependent contact phase.

We frame force-aware VLA as a modality grounding problem, where grounding denotes estimating the task-conditioned admissibility of physical feedback from the VLA's visual-language state before measured force is allowed to affect action generation.
Recent advances in multimodal foundation models have shown that heterogeneous inputs can be grounded in pretrained backbones through intermediate interfaces such as latent queries, resamplers, and gated attention~\citep{jaegle2021perceiverio,alayrac2022flamingo,li2023blip2}.
These interfaces transfer information into VLM representation spaces without requiring raw non-visual signals to be inserted as ordinary input tokens.
For force-aware VLA, such an interface can use visual-language context to assess when physical feedback should influence action generation, taking task progress and contact phase into account.

This perspective motivates separating the estimation of force relevance from the encoding of force measurements.
Such a separation calls for a mechanism that uses the VLA's visual-language task state to modulate the contribution of measured wrench histories to action generation.
FR-VLA instantiates this mechanism with force query states that infer a force admission signal from visual-language context, while measured wrench histories remain in a separate action-side force pathway.
The resulting admission behavior varies with task progress rather than simply with wrench availability, as illustrated in Figure~\ref{fig:teaser}(b).

Real-world evaluations on two robotic platforms and five contact-rich tasks show improved performance and robustness over vision-only and directly force-conditioned VLA baselines.
Our contributions are summarized as follows:
(i) Force-aware VLA is formulated as a modality grounding problem for low-dimensional physical feedback, where force relevance is inferred from visual-language task context rather than assumed from sensor availability.
(ii) FR-VLA introduces a force admission architecture in which visual-language states determine when force can influence action generation, while measured wrench histories are encoded separately for action-side control.
(iii) A detailed analysis of grounded force admission shows that the learned admission behavior follows task phase interaction regimes rather than instantaneous force magnitude.

\begin{figure}[!t]
  \centering
  \includegraphics[width=0.94\linewidth]{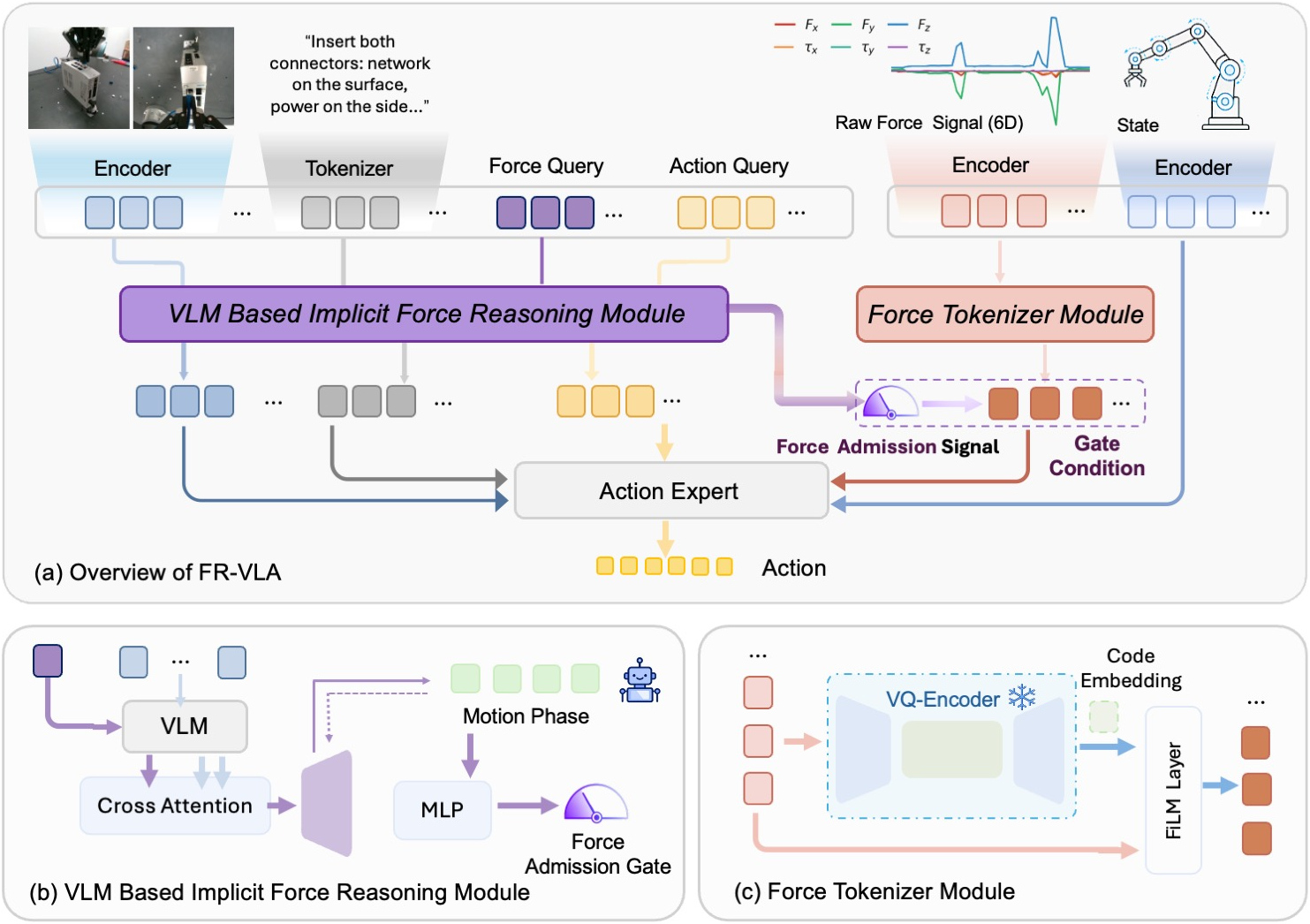}
	\caption{
	\textbf{Overview of FR-VLA.}
	{(a)} Overall architecture: four token groups are processed by the VLM-based Implicit Force Reasoning Module, which infers contact-relevant motion phases and produces a force admission signal.
	Measured wrench histories are encoded by the Force Tokenizer Module and enter the Action Expert only through the admission-conditioned action-side pathway.
	{(b)} The VLM-based Implicit Force Reasoning Module grounds force query states with visual features through cross-attention and decodes them into motion phase inference, producing the admission signal that determines when force information becomes action-relevant.
	{(c)} The Force Tokenizer Module discretizes wrench histories into contact-semantic codes via a frozen VQ-Encoder and uses their Code Embeddings to FiLM-condition the force pathway.
	}
  \label{fig:pipeline}
\end{figure}

\section{Related Work}
\label{sec:related_work}

\textbf{Force-Aware Policy Learning and Contact Control.}
Classical contact control provides model-based approaches to compliant manipulation.
Hybrid position/force control assigns position and force objectives to complementary directions in a task-related coordinate frame~\citep{raibert1981hybrid}.
Impedance control specifies the dynamic relationship between motion and interaction force, shaping the robot's mechanical response to its environment~\citep{hogan1985impedance}.
Force feedback has been studied in both policy learning and contact control~\citep{zhu2019flexure,zhu2022weightimprinting,xue2025rdp,ranjbar2021residual,zhang2023admittance}.
ForceMimic combines force-aware demonstration collection with imitation learning of force--motion commands and hybrid force-position control~\citep{liu2025forcemimic}.
Force Policy separates vision-guided global behavior from a local force feedback policy that estimates an interaction frame for hybrid force-position control~\citep{forcepolicy2026}.
Learning variable impedance provides another approach to adapting robot behavior under contact uncertainty~\citep{bogdanovic2019variableimpedance,martin2019vices}.
Beyond control design, effective multimodal learning requires preventing policies from neglecting force feedback.
FACTR addresses this issue through a curriculum that progressively reduces visual corruption, encouraging the policy to use force information rather than overfit to vision~\citep{factr2025}.
CRAFT similarly addresses modality imbalance in VLAs, using a variational information bottleneck to regulate visual-language representations during curriculum fine-tuning~\citep{craft2026}.
FR-VLA addresses force utilization through an explicit architectural separation between context-dependent force admission and wrench history representation.

\textbf{Force-Aware VLAs and Contact-Dependent Fusion.}
Recent approaches incorporate physical feedback into VLAs through different interfaces~\citep{tactilevla2025,favla2026,zhang2025vtla,cheng2025omnivtla}.
ForceVLA introduces a force-aware mixture of experts for action generation, while ForceVLA2 combines force-based prompts in the VLM with cross-scale fusion in the action expert for hybrid force-position control~\citep{forcevla2025,forcevla2_2026}.
Other approaches exploit physical information during training while reducing deployment-time sensing requirements: FD-VLA distills force information into latent tokens, and HapticVLA transfers tactile-aware behavior to a student that operates without inference-time tactile sensing~\citep{fdvla2026,hapticvla2026}.
Prediction-based designs anticipate future interaction: FoAR uses future contact prediction to guide visual--force fusion and dynamically adjust force use across contact and non-contact phases~\citep{foar2024}, and 3D world model policies forecast future scene states to guide action generation~\citep{wang2026dream}.
In FR-VLA, force query states infer interaction phases from visual-language context, and the resulting admission signal biases action-side attention to a separately represented wrench history.
This phase-supervised interface connects task context to force utilization while keeping measured wrench inputs in a dedicated action-side pathway.

\textbf{Cross-Modal Interfaces and Grounding.}
Latent queries and gated interfaces offer useful building blocks for connecting heterogeneous modalities.
Perceiver IO uses latent arrays to process structured inputs, Flamingo introduces gated cross-attention between visual features and a language model, and BLIP-2 uses a query-based interface between frozen visual and language models~\citep{jaegle2021perceiverio,alayrac2022flamingo,li2023blip2}.
FR-VLA adopts a task-specific query interface: its force query states receive motion phase supervision to ground force admission in visual-language context.
Measured wrench histories remain outside the VLM token stream and condition the action model through a separate force pathway.
This design distinguishes the inference of force relevance from the representation of measured physical feedback.

\section{FR-VLA Model}
\label{sec:method}

\subsection{Motivation}
\label{sec:failure_observation}

\begin{wrapfigure}[16]{r}{0.46\linewidth}
  \vspace{-12pt}
  \centering
  \includegraphics[width=\linewidth]{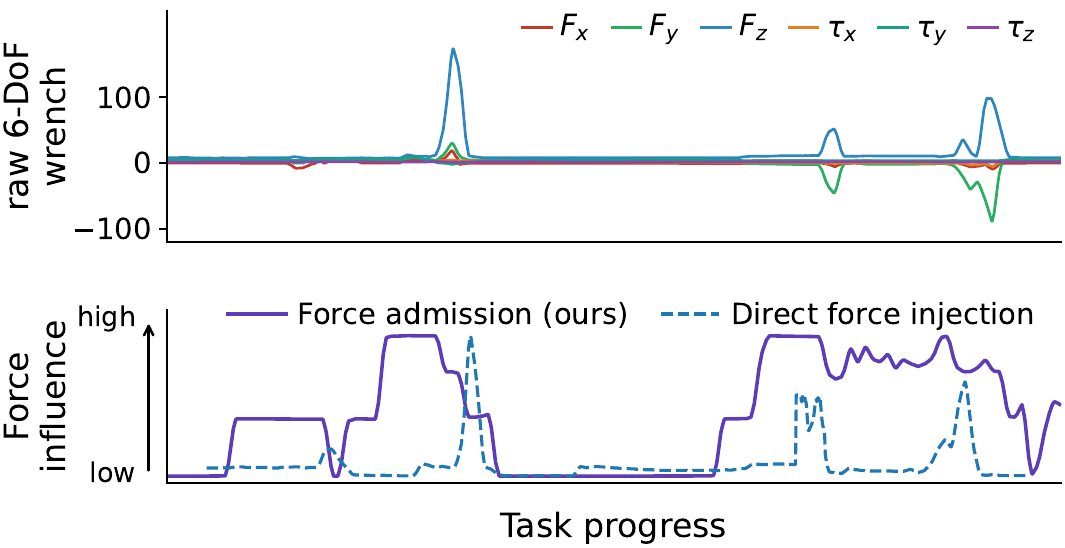}
  \captionsetup{skip=2pt}
  \caption{Trajectory-level comparison between raw wrench dynamics and learned force admission.
  Top: raw 6-DoF wrench during connector insertion.
  Bottom: force admission signal in FR-VLA \(g^f_t\) (solid) sustains force eligibility across contact-relevant windows, whereas a direct injection baseline (dashed) tracks wrench magnitude and spikes only at large transients.}
  \label{fig:force_admission}
\end{wrapfigure}

Direct force use exposes a behavioral ambiguity because making wrench measurements continuously available does not specify when force should become eligible to affect action generation.
Force often becomes informative over temporal regimes in which wrench dynamics evolve before and during contact changes, not only at instants of large measured transients.

This ambiguity is evident in precision insertion, where a large contact force may indicate constraint or misalignment, while a subsequent reduction in force during corrective motion remains decision relevant because it indicates that the adjustment is releasing the constraint.
A useful policy should therefore maintain force eligibility across such regimes, allowing the action model to observe evolving force patterns rather than react only to magnitude peaks, as shown in Figure~\ref{fig:force_admission}.
This distinction motivates treating force admission as a task-conditioned eligibility window over force dynamics.
FR-VLA therefore separates force relevance from force representation, as shown in Figure~\ref{fig:pipeline}: force query tokens estimate how strongly force should be admitted, and the Force Tokenizer represents the measured wrench on the action side.

\subsection{Implicit Force Reasoning Module}
\label{sec:force_reasoning}

Force query tokens provide a VLM-internal interface for force admission without carrying measured wrench values.
They are optimized to infer an interaction phase state from visual-language context, rather than to reconstruct force measurements or predict actions directly.
This allows the VLM to acquire force awareness while keeping raw physical feedback outside the token stream, as shown in Figure~\ref{fig:pipeline}(b).
The force query states \(h^f_t\) infer a motion phase distribution, from which a signed admission signal is derived as
\begin{equation}
    \hat{p}_t = \mathrm{softmax}\!\left(P_{\phi}(h^f_t)\right),
    \label{eq:phase-prediction}
\end{equation}
\begin{equation}
    g^f_t = G_{\eta}(\hat{p}_t)
    = s_{\mathrm{adm}}
    \left(1-\frac{\mathcal{H}(\hat{p}_t)}{\log K_p}\right)
    \tanh\!\left(M_g(\hat{p}_t)\right).
    \label{eq:reasoning-gate}
\end{equation}
Here, \(P_\phi\) is the phase inference head, \(M_g\) is an MLP, and \(K_p\) is the number of task-specific phases.
The \(\tanh\) output promotes or suppresses force attention, while the learnable positive scale \(s_{\mathrm{adm}}\) controls its magnitude; its parameterization is given in Appendix~\ref{app:gate-conditioning}.
Normalized entropy measures phase uncertainty; greater uncertainty attenuates the adjustment toward zero, returning the logit offset toward its baseline.

\noindent\textbf{Motion phase supervision.}
We generate phase labels offline from the robot's 6D wrench measurements and gripper states.
Gripper opening transitions partition demonstrations into manipulation segments.
Within each segment, a shared detector identifies contact events from force deviations relative to an adaptive reference estimated from recent non-contact frames, using hysteresis and temporal persistence.
Major events are selected based on both absolute and segment-relative peak deviations.
The contact interval extends from the first major event's onset to the final force release relative to its pre-contact reference, spanning intermediate retractions and brief force-free gaps; the remaining frames are labeled free.
For tasks with extended pre-contact approach periods, we further distinguish an approach phase.
These labels provide auxiliary supervision for phase prediction from visual-language context.
The full labeling procedure is provided in Appendix~\ref{app:phase-label-generation}.

\subsection{Force Tokenizer and Admission-Conditioned Attention}
\label{sec:force-vq}

\begin{figure}[!t]
  \centering
  \includegraphics[width=0.9\linewidth]{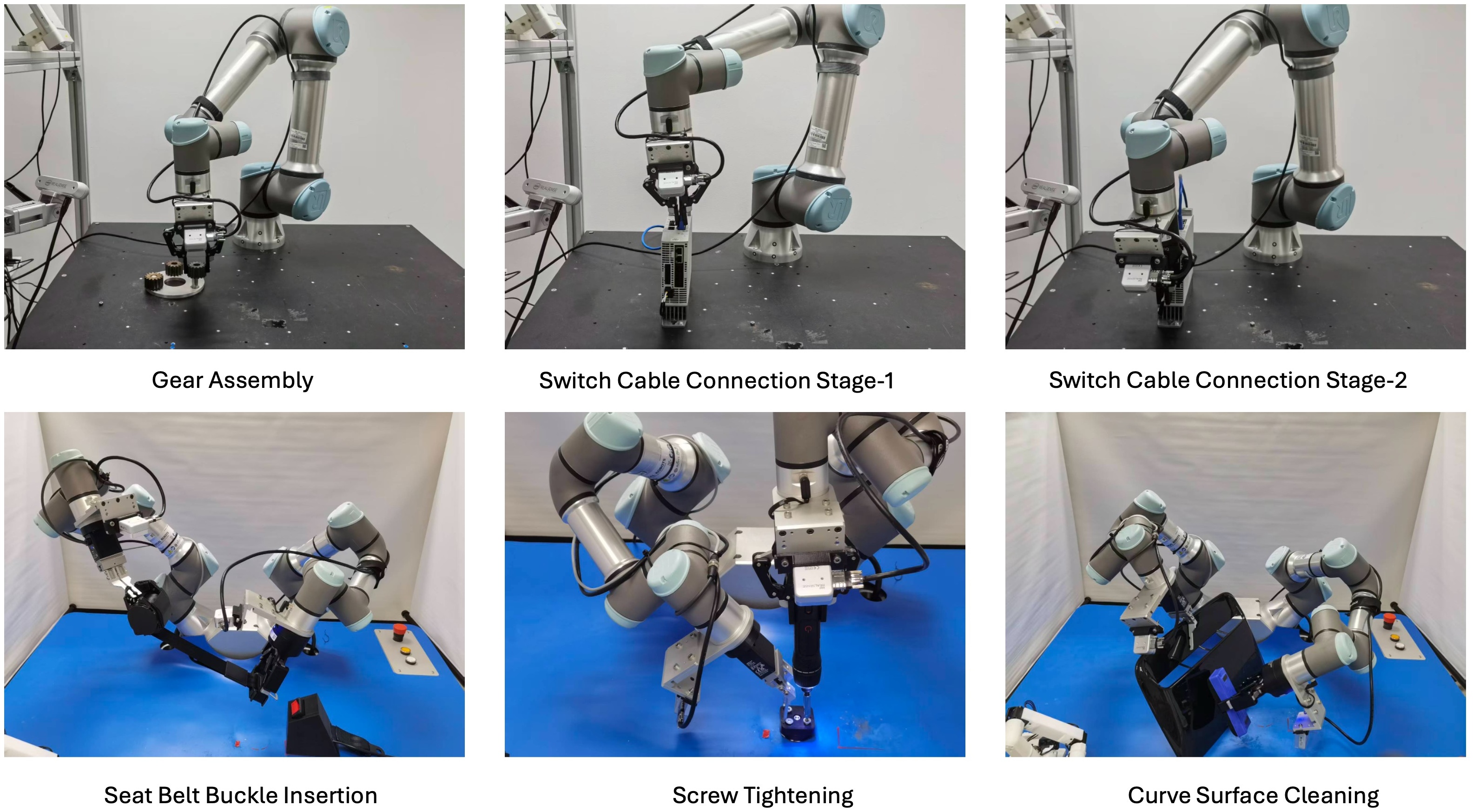}
  \captionsetup{skip=4pt}
\caption{
  \textbf{Real-world contact-rich manipulation benchmark.}
  The evaluation suite spans Gear Assembly, Switch Cable Connection with a visually occluded side power connector stage, Curved Surface Cleaning, Screw Tightening, and Seat Belt Buckle Insertion, covering contact transitions, constrained insertion, sustained surface interaction, and visually limited contact reasoning.}
  \label{fig:task-overview}
\end{figure}

The force admission signal determines how strongly force is promoted or suppressed in action generation, but the promoted force still requires a stable action-side representation.
Raw wrench histories are high-frequency and magnitude-sensitive; using them as continuous control features alone can make action generation overly responsive to transient contact changes.
The Force Tokenizer Module addresses this by assigning short wrench histories to recurring contact-dynamic patterns, so local wrench variations that correspond to the same contact regime share a stable code embedding.

The Force Tokenizer is implemented as a VQ-VAE~\citep{oord2017vqvae}, pretrained on demonstration wrench histories and frozen during VLA training.
Given a wrench history $W_t$, the VQ encoder selects the nearest codebook entry.
Its embedding does not replace the continuous wrench representation; instead, it generates FiLM~\citep{perez2017film} scale and shift parameters that modulate the raw force tokens, so the action model captures recurring contact patterns while retaining detailed wrench information, as illustrated in Figure~\ref{fig:pipeline}(c).
Implementation details are given in Appendix~\ref{app:tokenizer-impl}.

Force admission is implemented within an adapter-style action bridge following VLA-Adapter~\citep{wang2025vlaadapter}, where action query latents serve as the interface between VLM representations and action generation.
FR-VLA incorporates force information into this bridge through admission-conditioned attention.
Let \(S^{x,m}_t\) denote the scaled attention scores for source \(x\) in head \(m\), with the force scores defined as
\begin{equation}
    S^{w,m}_t = \frac{Q^{a,m}_t(K^{w,m}_t)^\top}{\sqrt{d_h}}.
    \label{eq:force-scaled-scores}
\end{equation}
The force admission signal \(g^f_t\) and a learned per-head bias \(\beta_m\) jointly shift all force token scores:
\begin{equation}
    \widetilde{S}^{w,m}_t = S^{w,m}_t + \beta_m + g^f_t.
    \label{eq:force-attention-scores}
\end{equation}
The scores from all sources are then concatenated and jointly normalized:
\begin{equation}
    A^m_t = \mathrm{softmax}\!\left(
    [S^{\mathrm{self},m}_t\mid S^{\mathrm{adapter},m}_t\mid
    S^{\mathrm{vision},m}_t\mid\widetilde{S}^{w,m}_t]
    \right).
    \label{eq:gated-force-admission}
\end{equation}
Here, \(Q^{a,m}_t\) and \(K^{w,m}_t\) are the action query and force key projections, and \(d_h\) is the head dimension.
The bias \(\beta_m\) captures each head's baseline preference for force information, while \(g^f_t\) provides a context-dependent adjustment.
This shared offset preserves relative scores among force tokens while modulating their total attention mass.

\subsection{Training Objective}
\label{sec:training}

Training proceeds in two stages.
First, the task-specific Force Tokenizer is pretrained from demonstration wrench histories and then frozen.
Second, FR-VLA is trained with behavior cloning and phase supervision.
The policy predicts an \(H\)-step action chunk \(\hat{A}_t=(\hat{a}_t,\ldots,\hat{a}_{t+H-1})\) from the current observation, matched to the demonstrated chunk \(A_t=(a_t,\ldots,a_{t+H-1})\).
The training objective is
\begin{equation}
    \mathcal{L}_{\mathrm{FR}}
    =
    \left\|\hat{A}_t - A_t\right\|_1
    +
    \lambda_p\,
    \mathrm{CE}\!\left(\hat{p}_t, y^p_t\right),
    \label{eq:training-loss}
\end{equation}
where the \(L_1\) term is summed over the action horizon, \(\hat{p}_t\) is the inferred motion phase distribution, and \(y^p_t\) is the offline phase label.
During training, phase inputs to downstream modules are detached: the phase loss trains the force query phase inference module, while the action loss trains the admission gate and action expert without propagating gradients to the phase logits.
At inference, the model infers the current contact phase from visual-language context, adjusts its attention to force accordingly, and generates the action chunk.

\section{Experimental Results}
\label{sec:result}

\begin{figure}[!t]
\centering
\includegraphics[width=0.8\linewidth]{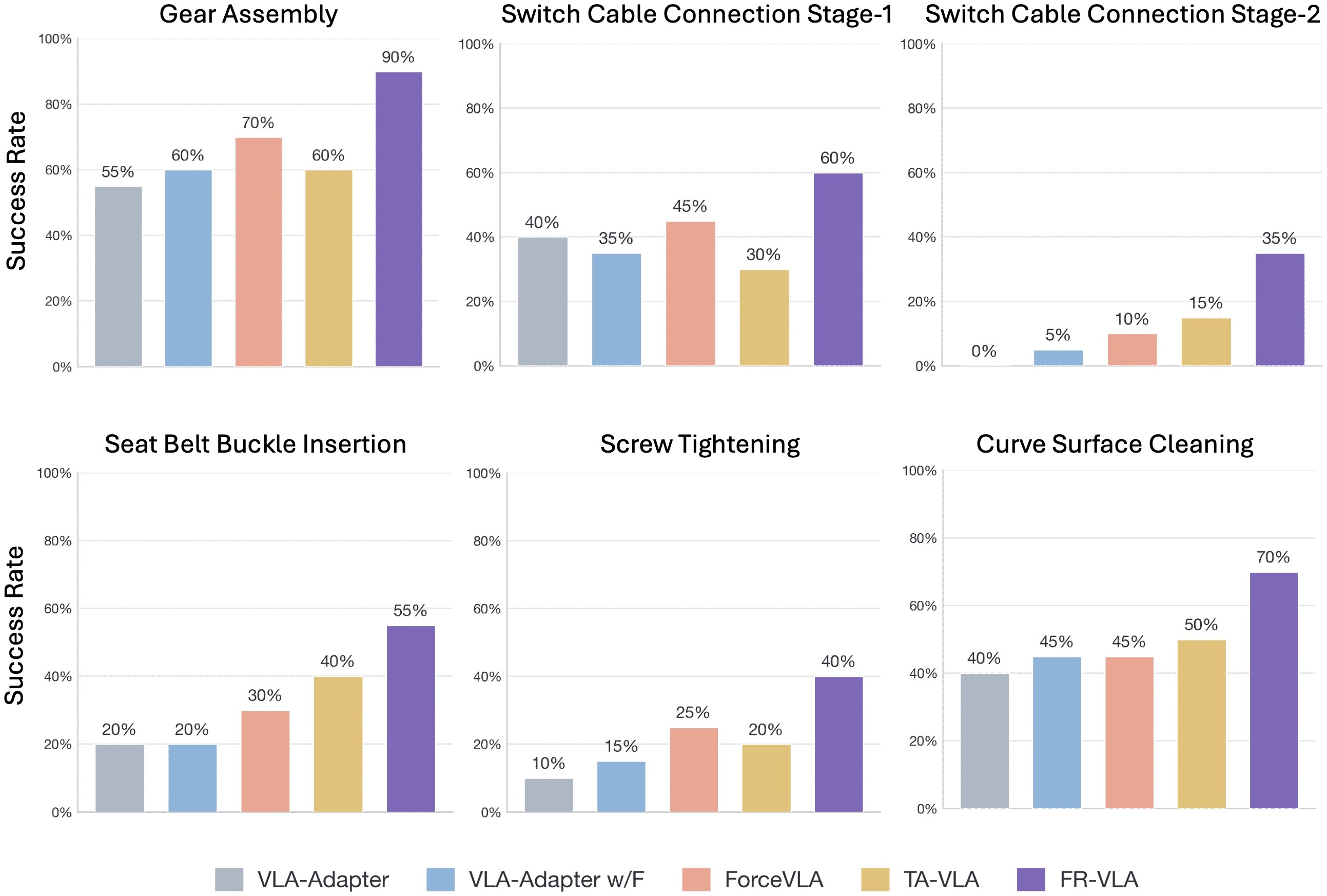}
\captionsetup{skip=4pt}
\caption{\textbf{Real-world success rates.}
Per-task/stage success rate over 20 trials for each method; FR-VLA (ours) is shown in purple.
Switch Cable Connection is evaluated in two stages, covering top Ethernet cable insertion and side power connector insertion.
VLA-Adapter w/F denotes directly inputting time-aligned 6D force together with robot state.}
\label{tab:main-results}
\end{figure}

Our experiments investigate four questions: (1) How does FR-VLA compare with other methods across different types of contact-rich manipulation tasks? (2) How does the learned force admission signal vary across interaction phases? (3) How do force admission and the Force Tokenizer contribute to performance? (4) How robust is FR-VLA to changes in objects and target positions?
We address these questions through real-world comparisons, trajectory analysis, component ablations, and perturbation evaluations.

\subsection{Experimental Setup}
\label{sec:exp-setup}

FR-VLA is evaluated on two Universal Robots platforms, UR5e and UR3e, using the robots' built-in 6D force/torque sensing as the wrench feedback source.
Each setup is observed by an Intel RealSense D455 third-view camera and a wrist-mounted Intel RealSense D405 camera.
We build a benchmark with 70--90 demonstrations per task, as shown in Figure~\ref{fig:task-overview}.
FR-VLA is compared against four baseline methods: VLA-Adapter~\citep{wang2025vlaadapter}, VLA-Adapter w/F, ForceVLA~\citep{forcevla2025} with force-aware MoE fusion during action decoding, and TA-VLA~\citep{zhang2025tavla} with decoder-side torque-action joint diffusion.
All methods use the same data and environmental conditions; detailed parameters are provided in Appendix~\ref{app:exp-params}.\looseness=-1

\subsection{Real-World Contact-Rich Manipulation Results}
\label{sec:exp-main}

As shown in Figure~\ref{tab:main-results}, FR-VLA achieves the highest success rate on all six evaluated stages across five tasks. Its mean success rate of 58.3\% exceeds that of the strongest baseline by 20.8 percentage points. The main improvement comes from stages where force is informative only under the right task context, rather than from simply adding wrench feedback to the policy. In Switch Cable Connection, visual evidence becomes unreliable during the side power connector insertion, making contact feedback essential once insertion begins. However, the wrench rises sharply at the moment of contact, and policies that only react passively to this force signal are prone to jamming during insertion without recovering. FR-VLA instead admits force according to the inferred interaction phase, allowing transient contact cues to be exploited when they are informative without continuously driving the policy. In Screw Tightening, the useful signal is not the absolute wrench magnitude, but the transition between alignment, engagement, and tightening, where small exploratory contacts and transient changes indicate whether the screw has entered the correct regime. These results suggest that FR-VLA improves performance by deciding when wrench dynamics should affect action generation.

We further compare FR-VLA with a simpler wrench magnitude gating baseline on Gear Assembly, which regulates force admission by the measured wrench amplitude without phase supervision.
FR-VLA achieves 90\% success, compared with 70\% for magnitude-based gating over 20 trials per method.
This indicates that wrench magnitude alone does not adequately reflect the relevance of force to action selection: both high forces during constrained insertion and decreasing forces during corrective motion can be informative.
A representative magnitude gating trajectory and further analysis are provided in Appendix~\ref{app:magnitude-gating}.\looseness=-1

\begin{figure}[!t]
  \centering
  \includegraphics[width=1.0\linewidth]{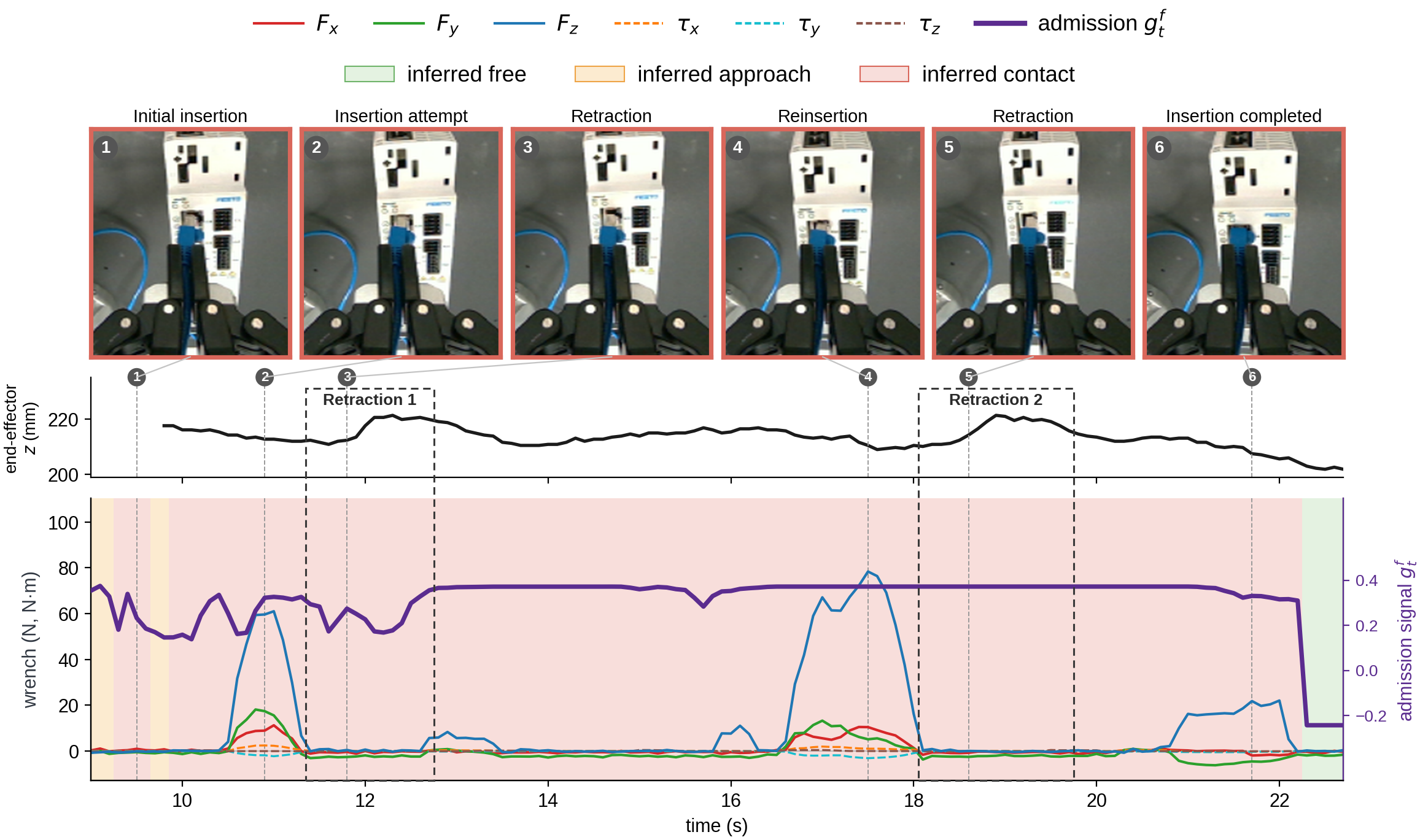}
  \captionsetup{skip=4pt}
\caption{
  \textbf{Force admission trajectory in Switch Cable Connection.}
  Keyframes, end-effector height, 6D wrench, and admission signal $g^f_t$ over a rollout with two retractions before insertion is completed.
  Shaded regions denote inferred phases; admission stays positive throughout the inferred contact phase and becomes negative once the phase switches to free.}
  \label{fig:gate-behavior}
\end{figure}

\subsection{Analysis of Force Admission Behavior}
\label{sec:exp-gate-analysis}

As shown in Figure~\ref{fig:gate-behavior}, the policy jams twice before completing the Ethernet cable insertion.
Each time the plug jams and the wrench rises sharply, the policy promptly retracts instead of continuing to push, and then reinserts.
During both retractions, the measured wrench returns close to zero, yet force admission remains positive, so attention to force is maintained throughout the recovery; admission becomes negative only after the insertion is completed and the inferred phase switches to free.
This shows that grounding force admission in the interaction phase keeps force feedback available throughout contact-rich recovery, allowing the model to actively exploit force signals.
Analyses for all five tasks are provided in Appendix~\ref{app:admission-across-tasks}.
The phase inference balanced accuracy reported in Table~\ref{tab:phase-accuracy} illustrates the grounding ability of FR-VLA, and the evaluation protocol is described in Appendix~\ref{app:phase-eval-protocol}.
Most phase prediction errors occur near contact onset and release, where gradual changes in visual observations may make phase boundaries difficult to identify.\looseness=-1

\begin{table}[!htb]
\centering

\caption{Phase inference accuracy.
Balanced accuracy after merging approach into free.
Switch Cable Connection averages its two stages.}
\label{tab:phase-accuracy}
\small
\setlength{\tabcolsep}{7pt}
\renewcommand{\arraystretch}{1.0}
\begin{tabular}{llr}
\toprule
Task group & Task & Balanced acc. \\
\midrule
Insertion & Gear Assembly & 93.7\% \\
Insertion & Switch Cable Connection & 90.5\% \\
Insertion & Seat Belt Buckle Insertion & 88.5\% \\
\midrule
Sustained contact & Curved Surface Cleaning & 80.6\% \\
Sustained contact & Screw Tightening & 86.2\% \\
\bottomrule
\end{tabular}
\end{table}

\subsection{Ablation Study}
\label{sec:exp-ablation}

We evaluate the contributions of force admission and the Force Tokenizer through component ablations, as shown in Table~\ref{tab:ablation-avg}.
Removing force admission reduces the mean success rate from 58.3\% to 35.0\%, while removing the Force Tokenizer reduces it to 40.0\%.
Both ablations lower success rates across all six evaluated stages, supporting the complementary roles of context-conditioned force attention and contact-dynamic representations.
To examine what the Force Tokenizer captures, we relate the selected codes to interaction events during contact.
Within each task, individual codes are associated with specific events such as jamming, holding, and withdrawing; in Screw Tightening, one code is 3.3 times more likely than average to co-occur with jamming, while another is 5.0 times more likely to co-occur with withdrawing.
The tokenizer therefore distinguishes interaction regimes within the contact phase rather than treating contact as a single state.
Further analysis of force code usage is provided in Appendix~\ref{app:ablation-study}.

\begin{table}[H]
\centering

\caption{Ablation results on force admission and force tokenization.}
\label{tab:ablation-avg}
\small
\setlength{\tabcolsep}{3.5pt}
\renewcommand{\arraystretch}{1.12}
\begin{tabular*}{\linewidth}{@{\extracolsep{\fill}}lrrrrrrr@{}}
\toprule
Variant & Gear & Switch-Top & Switch-Side & Cleaning & Screw & Buckle & Avg. \\
\midrule
w/o force admission & 60\% & 30\% & 15\% & 50\% & 25\% & 30\% & 35.0\% \\
w/o force tokenization & 70\% & 40\% & 20\% & 55\% & 20\% & 35\% & 40.0\% \\
\textbf{FR-VLA} & \textbf{90\%} & \textbf{60\%} & \textbf{35\%} & \textbf{70\%} & \textbf{40\%} & \textbf{55\%} & \textbf{58.3\%} \\
\bottomrule
\end{tabular*}
\end{table}

\subsection{Robustness to Object and Position Perturbations}
\label{sec:exp-robustness}

We evaluate robustness to object and position changes in Gear Assembly and Switch Cable Connection.
For Gear Assembly, we test two unseen gears and two groups of unseen target peg positions, varying contact geometry and alignment conditions, respectively.
For Switch Cable Connection, we test four groups of unseen initial Ethernet cable positions that alter the approach direction and contact angle during top port insertion, as illustrated in Appendix~\ref{app:robustness}.

As shown in Tables~\ref{tab:gear-robustness} and~\ref{tab:cable-robustness}, FR-VLA outperforms VLA-Adapter w/F by 25 percentage points under both gear and target position changes, and completes the insertion in half of the trials across the unseen cable positions.

\iflatexml\else
\noindent
\begin{minipage}[t]{0.48\linewidth}
\vspace{0pt}
\fi
\begin{table}[H]
  \centering
  \small
  \iflatexml\raggedright\fi
  \caption{Gear Assembly robustness over 20 trials per perturbation category.}
  \label{tab:gear-robustness}
  \setlength{\tabcolsep}{3pt}
  \renewcommand{\arraystretch}{1.12}
  \iflatexml
  \begin{tabular}{lcc}
  \else
  \begin{tabular*}{\linewidth}{@{\extracolsep{\fill}}lcc@{}}
  \fi
  \toprule
  Method & Unseen gear & Target perturb. \\
  \midrule
  VLA-Adapter w/F & 60\% & 50\% \\
  \textbf{FR-VLA} & \textbf{85\%} & \textbf{75\%} \\
  \bottomrule
  \iflatexml
  \end{tabular}
  \else
  \end{tabular*}
  \fi
\end{table}
\iflatexml
\par
\else
\end{minipage}\hfill
\begin{minipage}[t]{0.48\linewidth}
\vspace{0pt}
\fi
\begin{table}[H]
  \centering
  \small
  \iflatexml\raggedright\fi
  \caption{Switch Cable Connection under unseen initial cable positions, with 10 trials per group.}
  \label{tab:cable-robustness}
  \setlength{\tabcolsep}{3pt}
  \renewcommand{\arraystretch}{1.12}
  \iflatexml
  \begin{tabular}{cccc}
  \else
  \begin{tabular*}{\linewidth}{@{\extracolsep{\fill}}cccc@{}}
  \fi
  \toprule
  Group 1 & Group 2 & Group 3 & Group 4 \\
  \midrule
  4/10 & 6/10 & 5/10 & 5/10 \\
  \bottomrule
  \iflatexml
  \end{tabular}
  \else
  \end{tabular*}
  \fi
\end{table}
\iflatexml\else
\end{minipage}
\par\smallskip
\fi

Cable position changes are more challenging because the cable is deformable: changing its initial position alters its bending configuration and tension, and thus the forces transmitted to the connector during insertion.
The resulting wrench patterns are underrepresented in the demonstrations, making adaptation more demanding than compensating for rigid object pose changes.

\section{Conclusion}
\label{sec:conclusion}

This paper frames force integration in VLAs as a force grounding problem rather than a simple sensing or fusion problem.
FR-VLA separates force admission from force representation, using visual-language context to decide when wrench feedback should influence action generation while keeping measured force in an action-side pathway.
Real-world results show that this separation improves contact-rich manipulation and that learned force admission follows task phase interaction regimes rather than instantaneous force magnitude.\looseness=-1

\section{Limitations}

Although FR-VLA improves force admission and representation, fine-grained force-conditioned corrections remain challenging in precision manipulation.
Human demonstrations often use force changes to guide very small action adjustments, which can be diluted during behavior cloning because the corresponding action differences are small relative to the overall trajectory.
Future work should introduce more structured action-side mechanisms, such as force-sensitive residual heads or transition-focused objectives, to increase sensitivity to small but decision-critical force variations.
We will also evaluate, on larger-scale datasets, whether temporal changes across multiple visual frames can help anticipate more fine-grained force interaction states.\looseness=-1

\section*{Acknowledgments}
This research is supported by the National Robotics Programme (NRP) 2.0 funding initiative ``Domain-specific Robotics Foundation Models for Manufacturing (DS-RFM)''.

\newpage
\appendix
\renewcommand{\floatpagefraction}{0.5}
\section{Method Details}
\label{app:method-details}

\subsection{Admission-Conditioned Attention}
\label{app:gate-conditioning}

Eqs.~\ref{eq:phase-prediction}--\ref{eq:gated-force-admission} define phase
inference, admission, and joint attention. We use \(K_p=3\) for Gear Assembly
and Switch Cable Connection, and 2 otherwise, as shown in
Table~\ref{tab:app-phase-config}. The gate map is initialized near the
calibrated baseline, without an initial preference for promotion or
suppression. The positive scale in Eq.~\ref{eq:reasoning-gate} is
parameterized as
\begin{equation*}
    s_{\mathrm{adm}}=\alpha_g\exp(\sigma_{\mathrm{adm}}),
\end{equation*}
where \(\alpha_g\) is the multiplier and \(\sigma_{\mathrm{adm}}\) is the
learned log-scale. Before joint attention, force token features are normalized:
\begin{equation}
    \bar h^w_t=\mathrm{LN}(P_w(h^w_t)).
\end{equation}
The projector \(P_w\) and LN align force features with normalized
visual-language tokens.

\subsection{Force Tokenizer Implementation}
\label{app:tokenizer-impl}

The force representation uses two coupled but distinct pathways. A continuous
force history encoder produces the force tokens consumed by the action expert,
while a pretrained VQ tokenizer maps a recent wrench window to a
discrete contact-dynamic code.

The tokenizer operates on a history of \(L_{\tau}\) frames of 6-DoF wrench
measurements. Each window is flattened into a \(6L_{\tau}\)-dimensional vector and
standardized using per-dimension training set statistics. Let
\(W_{t-L_{\tau}+1:t}\in\mathbb{R}^{L_{\tau}\times6}\) denote the wrench window. The
tokenizer input is
\begin{equation}
    x^{\tau}_t =
    \frac{\mathrm{vec}(W_{t-L_{\tau}+1:t})-\mu_{\tau}}
    {\max(\sigma_{\tau},0.1)}\in\mathbb{R}^{6L_{\tau}},
    \label{eq:app-tactile-features}
\end{equation}
where division and the standard-deviation floor are applied elementwise, and
\(\mu_{\tau}\) and \(\sigma_{\tau}\) are computed from the training set.
The reconstruction target is this normalized \(6L_{\tau}\)-dimensional window.

The VQ module is trained separately from the VLA policy with a reconstruction
objective and a commitment term, while the codebook is maintained by an
exponential-moving-average update. During VLA training, the tokenizer is kept
frozen and is used only to provide the selected code embedding for each force
history.
The encoder maps \(x^{\tau}_t\) to a latent vector \(z_t=E_{\tau}(x^{\tau}_t)\).
Given a codebook \(\mathcal{C}_{\tau}=\{c^{\tau}_k\}_{k=1}^{K_{\tau}}\), the selected contact
code is
\begin{equation}
    k_t =
    \arg\min_k \|z_t-c^{\tau}_k\|_2,
    \qquad
    z^q_t =
    z_t + (c^{\tau}_{k_t}-z_t)_{\mathrm{sg}},
    \label{eq:app-vq-assign}
\end{equation}
where \((\cdot)_{\mathrm{sg}}\) denotes the stop-gradient straight-through
operation. The decoder reconstructs the normalized wrench window vector,
\(\hat{x}^{\tau}_t=D_{\tau}(z^q_t)\), and the tokenizer is pretrained with
\begin{equation}
    \mathcal{L}_{\tau}
    =
    \| \hat{x}^{\tau}_t - x^{\tau}_t \|_2^2
    +
    \lambda_{\mathrm{com}}
    \| z_t - (c^{\tau}_{k_t})_{\mathrm{sg}} \|_2^2.
    \label{eq:app-vq-loss}
\end{equation}
The encoder, decoder, and codebook configurations are given in
Appendix~\ref{app:model-training-params}.

The selected code embedding conditions the continuous force token through FiLM
modulation: a weighted code embedding predicts scale and shift terms that
are applied to the projected and normalized force history token. The FiLM projection is a linear layer with zero-initialized weights and bias,
so the initial FiLM transformation is the identity. The action expert initially
receives the same continuous force representation and gradually learns to use
the contact-dynamic code as a regime-dependent adjustment. This keeps fine
wrench details in the continuous token while using the discrete code to provide
a stable contact regime context.
Let \(r_t=F_w(W_{t-L+1:t})\) be the continuous force history representation and
let \(\bar{r}_t=\mathrm{LN}(P_w(r_t))\) be its projected token. The selected code
embedding is scaled as \(e_t=w_t c^{\tau}_{k_t}\), where \(w_t\) is a scalar
weight on the code embedding.
The FiLM parameters are then computed as
\begin{equation}
    [\gamma_t,\beta_t] = \mathrm{Linear}_{\mathrm{film}}(e_t),
    \qquad
    h^w_t =
    \bar{r}_t \odot (1+s_{\mathrm{film}}\gamma_t)
    + s_{\mathrm{film}}\beta_t .
    \label{eq:app-film}
\end{equation}
We use \(s_{\mathrm{film}}=1\), with no nonlinear activation applied to the predicted FiLM parameters.
The resulting \(h^w_t\) is the force token passed to the action bridge and then
regulated by the source-level admission mechanism in
Eqs.~\ref{eq:force-attention-scores}--\ref{eq:gated-force-admission}.

\subsection{Offline Motion Phase Label Generation}
\label{app:phase-label-generation}

Phase labels are generated offline from demonstration force measurements and
gripper states using a contact detection procedure shared across tasks.
Task-specific configurations and parameter values are given in
Appendix~\ref{app:phase-config}.

\paragraph{Inputs and manipulation segments.}
The detector uses the raw three-dimensional force \(F_t\in\mathbb{R}^3\)
from the robot's force/torque sensor, with no preliminary baseline
subtraction. Gripper states are used to partition
trajectories into manipulation segments. A transition of the gripper from
closed to open ends a segment; closed corresponds to a raw reading greater than
\(q_{\mathrm{raw}}\) or a normalized reading greater than
\(q_{\mathrm{norm}}\). A trajectory with unchanged
gripper states forms one segment. Gripper values do not enter the
contact event decision. The three-dimensional force used for labeling is
distinct from the six-dimensional wrench input to the Force Tokenizer.

\paragraph{Adaptive reference and event detection.}
At a sampling rate \(f_s\), the free motion reference \(F_{\mathrm{ref}}\)
is the componentwise median of the most recent \(N_{\mathrm{ref}}\)
non-contact force samples. The detector computes
\begin{equation}
    d_t=\|F_t-F_{\mathrm{ref}}\|_2.
    \label{eq:app-force-dynamics-label}
\end{equation}
An event starts when \(d_t\geq\theta_{\mathrm{on}}\) for
\(N_{\mathrm{on}}\) consecutive frames and ends when
\(d_t<\theta_{\mathrm{off}}\) for \(N_{\mathrm{off}}\) consecutive frames.
The reference is held fixed during contact, except for a steady load update:
when the standard deviation of \(\|F_t\|_2\) over the most recent
\(N_{\mathrm{steady}}\) frames is below \(\sigma_{\mathrm{steady}}\),
the stable interval is incorporated into the reference.
This update accommodates persistent loads such as cable or seat belt tension.
All event detection parameters are shared across tasks.

\paragraph{Major event selection and contact intervals.}
Within each manipulation segment, let \(m_j\) denote the peak force deviation
of event \(j\), and let \(m_{\max}\) be the largest event peak in that segment.
An event is retained as a major event when
\begin{equation}
    m_j\geq\theta_{\mathrm{major}},\qquad m_j\geq\rho m_{\max}.
\end{equation}
The contact label begins at the onset of the first major event. Its endpoint is
the last frame in the segment whose force deviation from that event's
pre-contact reference is at least \(\theta_{\mathrm{off}}\). The resulting interval includes
intermediate retractions, jamming, and brief force-free gaps. Thus, individual
event boundaries and the full contact label interval serve different roles.

\paragraph{Phase assignment.}
Frames within the resulting interval receive the contact label; the remaining
frames receive the free label. For tasks with an approach phase,
the \(N_{\mathrm{approach}}\) frames immediately preceding contact are instead labeled approach,
clipped to the current manipulation segment. These tasks use labels 0, 1, and 2
for free, approach, and contact, respectively. The other tasks use free and
contact as their two training categories. The phase head is supervised with
these task-specific categories and predicts phases from visual-language context
at inference time.

\section{Experimental Details and Analysis}
\label{app:experiments}

\subsection{Experimental Setup}
\label{app:exp-params}

\subsubsection{Task Setup}

\begin{figure}[t]
    \centering
    \includegraphics[width=\linewidth]{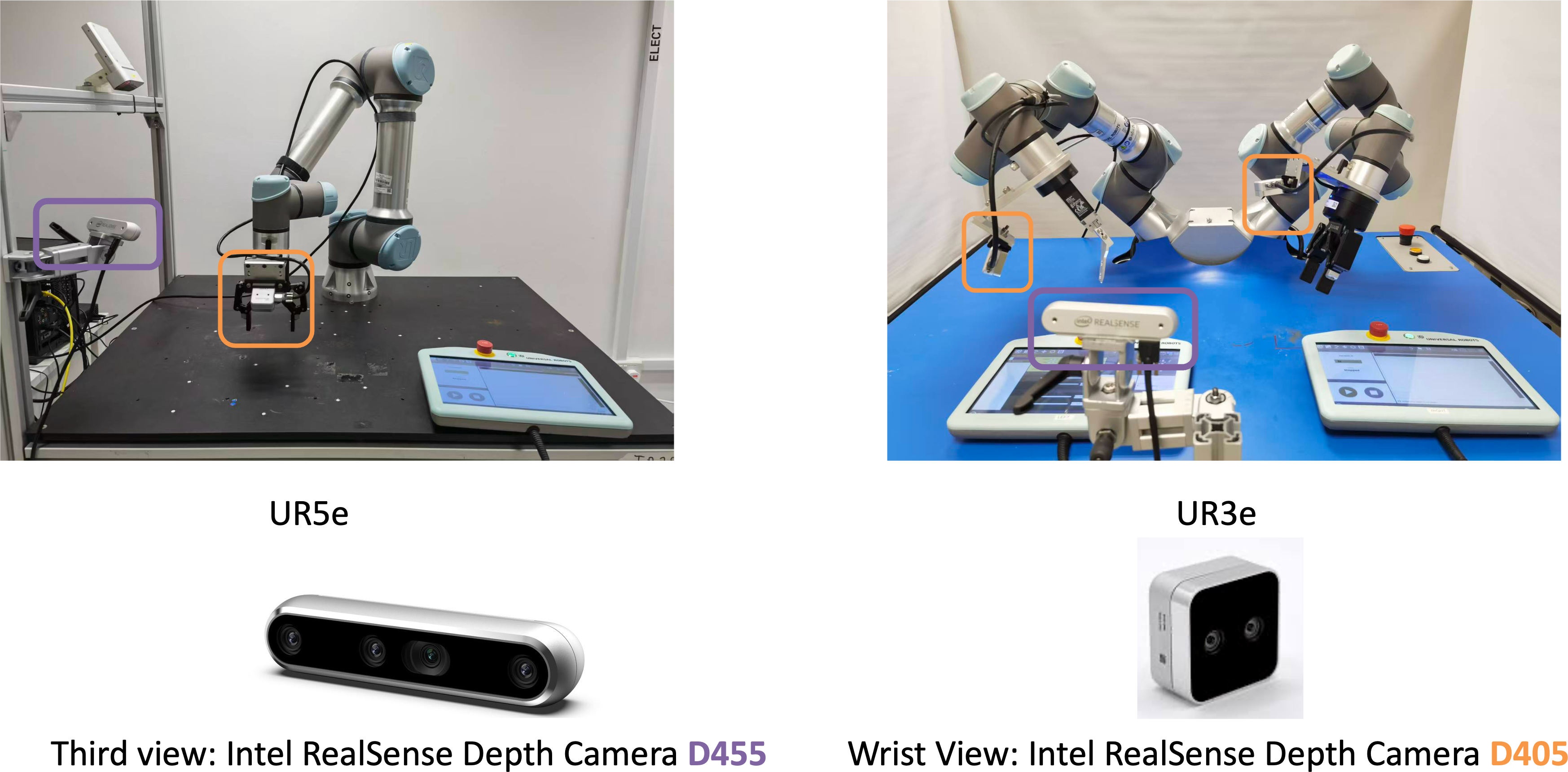}
    \caption{
    \textbf{Robot platforms and camera setup.}
    The two evaluation platforms, UR5e (left) and UR3e (right), each use the
    robot's built-in 6D force/torque sensing. Both share the same camera
    configuration: a third-view Intel RealSense D455 camera and a wrist-mounted
    Intel RealSense D405 camera, highlighted on each platform.}
    \label{fig:app-setup}
\end{figure}

We evaluate FR-VLA on five contact-rich manipulation tasks, using the
platforms shown in Figure~\ref{fig:app-setup}: Gear Assembly,
Switch Cable Connection, Curved Surface Cleaning, Screw Tightening, and Seat
Belt Buckle Insertion. Demonstrations are collected with a 3D mouse interface.
The operator commands the desired end-effector motion, while the robot executes
the command through a Cartesian impedance controller rather than a stiff
position-tracking controller. In end-effector coordinates, the commanded
interaction behavior can be written as
\begin{equation}
    \tau =
    J(q)^{\top}
    \left[
    K_x(x_d-x) + D_x(\dot{x}_d-\dot{x})
    \right]
    + \tau_{\mathrm{bias}},
    \label{eq:app-impedance-control}
\end{equation}
where \(x_d\) is the 3D-mouse-driven target pose, \(x\) is the measured
end-effector pose, \(J(q)\) is the robot Jacobian, and \(K_x,D_x\) define the
Cartesian impedance. This compliant data collection mode encourages smooth
contact behavior and reduces the effect of high-frequency human teleoperation
noise, which is especially important in precision contact tasks where abrupt
micro-motions can trigger protective stops.
During data collection, images from both cameras are recorded at 10\,Hz, while
force, robot state, and actions are recorded at 50\,Hz; for training, the force,
state, and action streams are aligned to the image timestamps.

Gear Assembly requires the robot to align a gear with a constrained peg and
complete insertion after contact. The task stresses the transition from visual
approach to force-guided alignment, where small changes in wrench dynamics
indicate whether the gear is seated correctly.

Switch Cable Connection contains two evaluated stages. In the top
Ethernet cable insertion stage, the target port location is varied and the robot
must localize the changed receptacle before insertion. In the side
power connector insertion stage, the robot inserts a plug into a lower side
connector where visual evidence is partially occluded and contact feedback is
needed once the plug begins to engage.

Curved Surface Cleaning requires sustained contact while following a curved
surface. Unlike insertion tasks, the informative force signal is not only the
first contact event; the policy must maintain contact without overreacting to
stable baseline forces.

Screw Tightening requires alignment, engagement, and rotational tightening. The
task depends on recognizing the transition between light exploratory contact and
successful thread engagement before applying the tightening motion.

Seat Belt Buckle Insertion requires pushing the buckle tongue into a latch
under limited visual observability near the final engagement point. The task
combines constrained insertion with a short contact transition at successful
latch engagement.

In all tasks, a trial is counted as successful only when the specified stage
objective is completed without human intervention.

\subsubsection{Model and Training Parameters}
\label{app:model-training-params}

FR-VLA predicts action chunks of $H=8$ steps, where each action is a 7-dimensional absolute end-effector pose with a gripper command, and the action expert receives a wrench history of $L=16$ frames.
The phase loss weight in Eq.~\ref{eq:training-loss} is $\lambda_p=0.5$.
The Force Tokenizer uses lightweight MLP encoder and decoder networks with a
codebook of \(K_{\tau}=8\) code embeddings.
Its input is a flattened window of \(L_{\tau}=8\) frames of 6D wrench measurements across all tasks. The tokenizer is pretrained per task on that
task's demonstration wrench histories and then frozen, while the action
interface and phase loss weight are shared across tasks. Phase head category counts are specified in Table~\ref{tab:app-phase-config}.
At inference time, the frozen tokenizer and FiLM conditioning run as part of
the policy forward pass, whereas the segmentation and contact event detection
used for phase labeling are applied only offline.
Training uses two NVIDIA RTX A6000 GPUs, and inference runs on a single NVIDIA RTX 3090 GPU.

\subsection{Wrench Magnitude Gating Baseline}
\label{app:magnitude-gating}

\begin{figure}[H]
    \centering
    \includegraphics[width=0.6\linewidth]{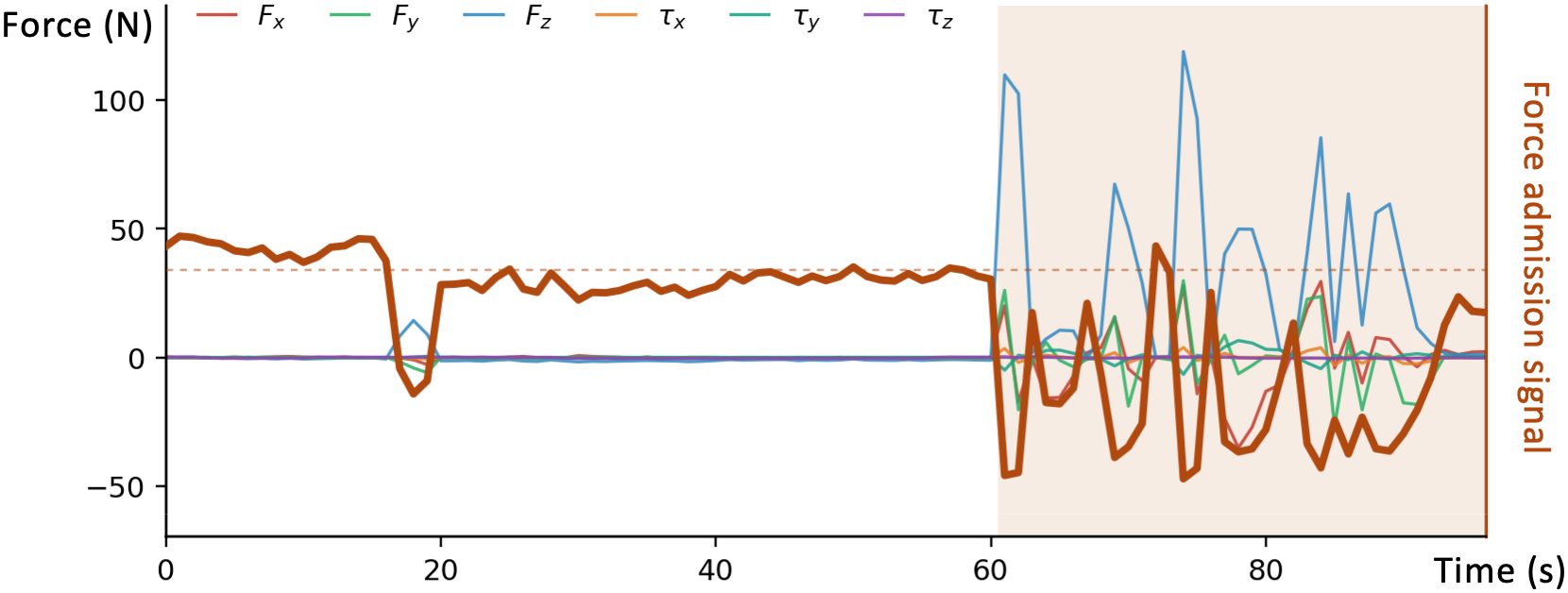}
    \captionsetup{skip=4pt}
    \caption{\textbf{Wrench magnitude gating on Gear Assembly.}
    Six-dimensional wrench measurements and the admission signal of the
    magnitude gating baseline during a representative execution.
    The admission curve varies with the measured wrench amplitude.}
    \label{fig:app-magnitude-gating}
\end{figure}

Figure~\ref{fig:app-magnitude-gating} shows a simpler wrench magnitude gating
baseline on Gear Assembly, in which the admission signal is computed from the
measured wrench amplitude without phase supervision. In this design, the
influence of force is determined by the measured wrench itself: it rises only
after a large transient has occurred and falls back once the magnitude
decreases. The policy can therefore only react passively to force feedback after
contact events, and cannot adjust its reliance on force in anticipation of an
upcoming interaction or maintain it when decision-relevant changes have small
magnitude, such as the force reduction during corrective motion. In contrast,
FR-VLA infers the interaction phase from visual-language context and uses it to
set the admission signal. Force eligibility can thus be raised as the task enters
a contact-relevant phase, sustained across the whole window, including intervals
where the wrench magnitude is small, and suppressed once the phase no longer
requires physical feedback. This allows FR-VLA to actively regulate how strongly
force influences action generation, rather than only responding to the
magnitude of the current measurement.

\subsection{Analysis of Force Admission Behavior}
\label{app:force-admission-analysis}

\begin{figure}[p]
    \centering
    \includegraphics[width=\linewidth]{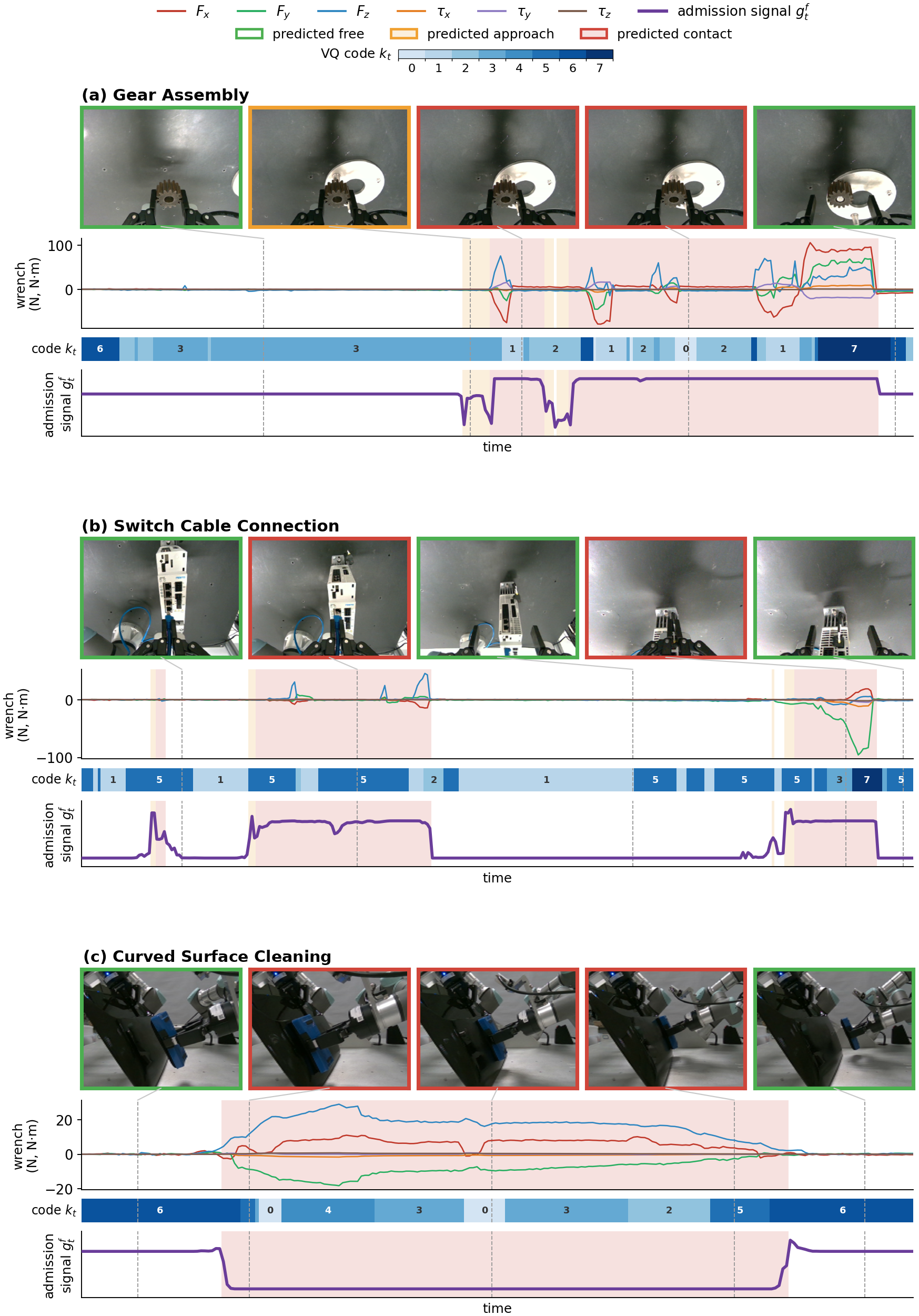}
    \captionsetup{skip=4pt}
    \caption{\textbf{Phase inference, force codes, and force admission across tasks.}
    Frame borders and shaded regions denote predicted phases, and the code strip shows the selected VQ code $k_t$.
    Additional trajectories are shown in Figure~\ref{fig:app-gate-all-tasks-more}.}
    \label{fig:app-gate-all-tasks}
\end{figure}

\begin{figure}[!t]
    \centering
    \includegraphics[width=\linewidth]{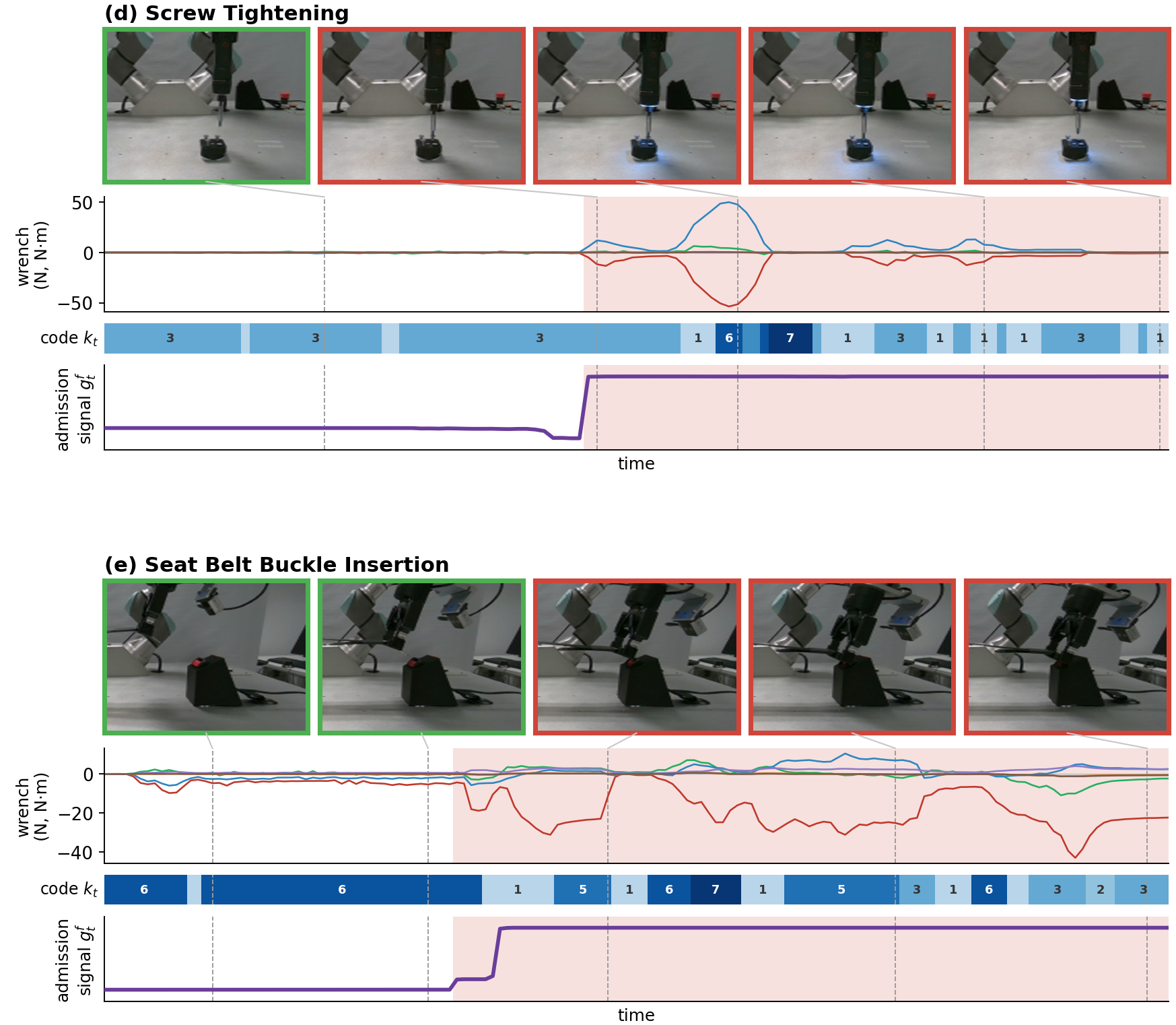}
    \captionsetup{skip=4pt}
    \caption{\textbf{Additional phase inference, force codes, and force admission trajectories.}
    Continuation of Figure~\ref{fig:app-gate-all-tasks}; frame borders, shaded regions, and the code strip use the same conventions.}
    \label{fig:app-gate-all-tasks-more}
\end{figure}

\subsubsection{Phase Supervision Configuration}
\label{app:phase-config}

Gear Assembly and Switch Cable Connection use three training classes (free,
approach, and contact), while the remaining tasks use two (free and contact), as
shown in Table~\ref{tab:app-phase-config}. Switch Cable Connection contains two
manipulation segments; each remaining task contains one.

\paragraph{Hyperparameter settings.}
We use \(f_s=10\,\mathrm{Hz}\), \(N_{\mathrm{ref}}=10\),
\(\theta_{\mathrm{on}}=5\,\mathrm{N}\), and
\(\theta_{\mathrm{off}}=2.5\,\mathrm{N}\).
The lower release threshold provides hysteresis, while
\(N_{\mathrm{on}}=3\) and \(N_{\mathrm{off}}=5\) require persistence
for 0.3\,s and 0.5\,s, respectively, to suppress short-lived fluctuations.
The steady load update uses \(N_{\mathrm{steady}}=15\) (1.5\,s) and
\(\sigma_{\mathrm{steady}}=1.5\,\mathrm{N}\).
Major events use \(\theta_{\mathrm{major}}=8\,\mathrm{N}\) and
\(\rho=0.5\), combining a minimum peak deviation with a segment-relative
criterion. The approach window is \(N_{\mathrm{approach}}=10\) (1\,s).
Gripper thresholds are \(q_{\mathrm{raw}}=100\) and
\(q_{\mathrm{norm}}=0.5\). These settings are fixed across tasks;
only the phase vocabulary differs, as shown in
Table~\ref{tab:app-phase-config}.

\begin{table}[H]
    \centering
    \small
    \setlength{\tabcolsep}{5pt}
    \caption{Phase supervision configuration. Contact detection uses
    shared thresholds; the training phase vocabulary differs by task.}
    \label{tab:app-phase-config}
    \begin{tabular}{lcl}
        \toprule
        Task & Manipulation segments & Training phases \\
        \midrule
        Gear Assembly & 1 & Free, approach, contact \\
        Switch Cable Connection & 2 & Free, approach, contact \\
        Curved Surface Cleaning & 1 & Free, contact \\
        Screw Tightening & 1 & Free, contact \\
        Seat Belt Buckle Insertion & 1 & Free, contact \\
        \bottomrule
    \end{tabular}
\end{table}

Table~\ref{tab:app-phase-label-distribution} reports the training label
distribution after merging approach into free. Switch Cable Connection is
reported as one task, aggregating its two manipulation segments.

\begin{table}[H]
    \centering
    \setlength{\tabcolsep}{8pt}
    \renewcommand{\arraystretch}{1.12}
    \caption{Distribution of offline motion phase labels.
    Relative prevalence of training labels after merging approach into free.
    Free includes approach; contact is unchanged. Switch Cable Connection is reported as one task rather than
    separating its two execution stages.}
    \label{tab:app-phase-label-distribution}
    \begin{tabular}{lrr}
        \toprule
        Task & Free & Contact \\
        \midrule
        Gear Assembly & 64\% & 36\% \\
        Switch Cable Connection & 53\% & 47\% \\
        Curved Surface Cleaning & 44\% & 56\% \\
        Screw Tightening & 56\% & 44\% \\
        Seat Belt Buckle Insertion & 63\% & 37\% \\
        \bottomrule
    \end{tabular}
\end{table}

\subsubsection{Evaluation Protocol for Phase Inference}
\label{app:phase-eval-protocol}

During each rollout, the phases predicted by the policy from visual-language
context are logged online together with the recorded observations, force
measurements, and gripper states. Ground-truth labels are then obtained post hoc
by applying the offline labeling procedure in
Appendix~\ref{app:phase-label-generation} to the recorded rollout data, and
phase inference accuracy is computed by comparing the logged predictions with
these labels.

For consistent reporting across tasks, we merge approach into free when
computing phase label distributions and phase inference accuracy. For
three-class models, the predicted free and approach probabilities are summed
to form the non-contact probability before computing binary balanced accuracy.
Specifically, \(\hat p_{\mathrm{free}}^{\mathrm{eval}}=
\hat p_{\mathrm{free}}+\hat p_{\mathrm{approach}}\) and
\(\hat p_{\mathrm{contact}}^{\mathrm{eval}}=\hat p_{\mathrm{contact}}\).
The larger of these two probabilities determines the predicted evaluation class;
ground-truth approach labels are mapped to free. Balanced accuracy is the mean
recall of the resulting free and contact classes. This aggregation is used for
reporting; phase supervision and gating retain the task-specific phase vocabulary.

\subsubsection{Force Admission across Tasks}
\label{app:admission-across-tasks}

Figures~\ref{fig:app-gate-all-tasks} and~\ref{fig:app-gate-all-tasks-more} show the predicted phases, wrench measurements, selected force codes, and force admission signals on all five tasks.
In Gear Assembly, Screw Tightening, and Seat Belt Buckle Insertion, admission rises at contact onset and remains high throughout the predicted contact phase, through alignment, engagement, and seating, whether the measured wrench is large or close to zero.
In Gear Assembly, where the rollout ends in free motion, admission returns to a lower level once the phase switches back to free.
Switch Cable Connection shows the same pattern in each of its two stages, with admission dropping during the transfer between the Ethernet cable and power connector insertions.
Curved Surface Cleaning exhibits the opposite behavior: admission is suppressed throughout the long period of stable contact and recovers once contact is released, since the steady wrench confirms ongoing contact but carries little new information for action selection.
Across tasks, changes in admission coincide with predicted phase boundaries rather than with wrench peaks, while the selected code typically remains stable during free motion and switches more often during contact, reflecting the distinct contact regimes encountered during interaction.

\subsection{Force Code Usage}
\label{app:ablation-study}
\label{app:code-analysis}

\begin{figure}[!tb]
    \centering
    \includegraphics[width=\linewidth]{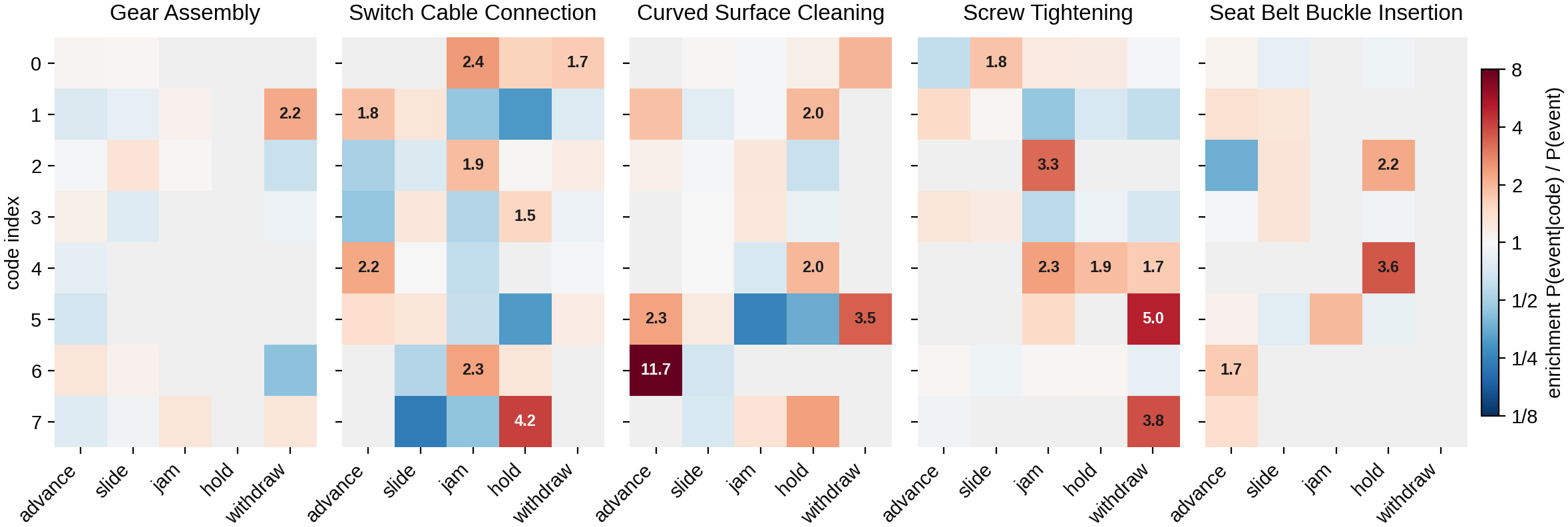}
    \captionsetup{skip=4pt}
    \caption{\textbf{Association between force codes and interaction events.}
    Each cell shows the lift $P(\mathrm{event}\mid\mathrm{code})/P(\mathrm{event})$ on a logarithmic color scale; red indicates enrichment and blue indicates depletion.}
    \label{fig:app-code-distribution}
\end{figure}

Figure~\ref{fig:app-code-distribution} relates the selected force codes to interaction events in each task.
The events describe end-effector motion during contact: advance moves along the insertion direction; slide moves tangentially under load, as in wiping or alignment; jam pushes under high load without progressing and is followed by a retraction; hold stays nearly still under high load, as when pressing, tightening, or seating; and withdraw retracts while contact is maintained.
Each cell reports the lift $P(\mathrm{event}\mid\mathrm{code})/P(\mathrm{event})$: values above 1 indicate that the event occurs more often than average when the code is selected, and values below 1 indicate that it occurs less often.

Because the Force Tokenizer is trained separately for each task, the same code index carries a task-specific meaning.
Within each task, individual codes are concentrated on particular interaction events.
In Switch Cable Connection, code 7 is enriched for hold and depleted for slide.
In Screw Tightening, code 2 is enriched for jam, whereas codes 5 and 7 are enriched for withdraw.
In Seat Belt Buckle Insertion, code 4 is enriched for hold, and in Curved Surface Cleaning, code 6 is strongly enriched for advance.
The association is weaker in Gear Assembly, where code 1 is mainly associated with withdraw.
Codes therefore distinguish interaction events within the contact phase, such as jamming, holding, and withdrawing, rather than representing contact as a single state.

\subsection{Robustness to Object and Position Perturbations}
\label{app:robustness}

\begin{figure}[H]
    \centering
    \includegraphics[width=0.5\linewidth]{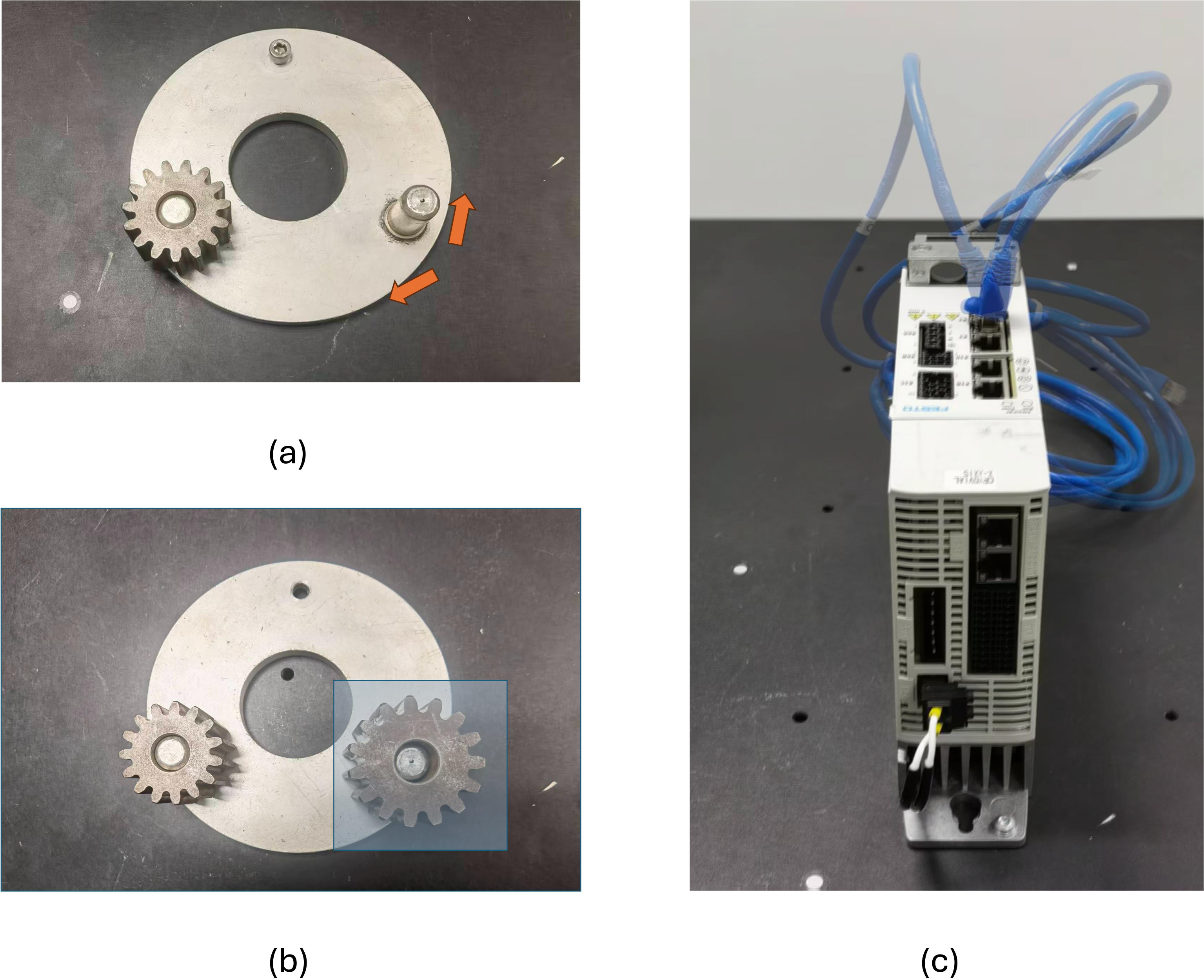}
    \captionsetup{skip=4pt}
    \caption{\textbf{Additional robustness test settings.}
    (a) Target peg position changes. (b) Unseen gears.
    (c) Initial Ethernet cable position changes.}
    \label{fig:app-robustness}
\end{figure}

Figure~\ref{fig:app-robustness} illustrates the perturbation settings.
Figure~\ref{fig:app-robustness}(a) shows the two groups of unseen target peg positions in Gear Assembly, which shift where the gear must be aligned before contact-guided insertion.
Figure~\ref{fig:app-robustness}(b) shows the two unseen gears, whose different geometry changes the contact conditions and shifts the force baseline during insertion.
Figure~\ref{fig:app-robustness}(c) shows the four groups of unseen initial Ethernet cable positions in Switch Cable Connection, which change the approach direction and contact angle during top port insertion.

These perturbations affect force feedback in different ways.
Unseen gears mainly alter the wrench observed during engagement and seating, so a policy that conditions directly on raw wrench values encounters force patterns outside its training distribution.
Target position changes mainly affect the visual approach and the timing of first contact.
Because admission is tied to the inferred phase rather than to a fixed point in the trajectory, force is still promoted once contact begins at the new location.
Cable position changes alter both the approach geometry and, through the deformable cable, the forces transmitted to the connector, producing visual and force conditions not seen during training.
FR-VLA succeeds in every cable position group, demonstrating that phase-grounded force admission generalizes to unseen initial configurations.
Although the cable configuration changes the appearance of the approach as well as the magnitude and direction of contact forces, the transition from approach to plug--port contact remains recognizable from visual-language context.
The policy therefore still promotes force at contact onset and uses the measured wrench to complete the final alignment and insertion under contact conditions that differ from the demonstrations.

\end{document}